 \documentclass[pmlr]{jmlr} 

\usepackage{microtype}
\usepackage{dirtytalk}
\usepackage{hyperref}
\usepackage{flushend}
\usepackage[justification=centering]{caption}
\usepackage{stfloats}
\usepackage{float}

\makeatletter
\newcommand*{\centerfloat}{%
  \parindent \z@
  \leftskip \z@ \@plus 1fil \@minus \textwidth
  \rightskip\leftskip
  \parfillskip \z@skip}
\makeatother

\def\hyphenateAndTtWholeString #1{\xHyphenate#1$\wholeString\unskip}

\def\xHyphenate#1#2\wholeString {\if#1$%
    \else\transform{#1}%
    \takeTheRest#2\ofTheString\fi}

\def\takeTheRest#1\ofTheString\fi
{\fi \xHyphenate#1\wholeString}

\def\transform#1{\url{#1}\hskip 0pt plus 1pt}

\def\urlx #1{\href{#1}{\hyphenateAndTtWholeString{#1}}}

\usepackage{booktabs}

\usepackage[load-configurations=version-1]{siunitx} 

\theorembodyfont{\upshape}
\theoremheaderfont{\scshape}
\theorempostheader{:}
\theoremsep{\newline}

\jmlryear{2021}
\jmlrworkshop{2nd Workshop on Diversity in Artificial Intelligence (AIDBEI)} 

\title[Characterizing Intersectional Group Fairness with Worst-Case Comparisons]{Characterizing Intersectional Group Fairness with Worst-Case Comparisons}

\author{%
\Name{Avijit Ghosh} \Email{ghosh.a@northeastern.edu}\\
\addr Fiddler Labs\thanks{Work done as a research intern at Fiddler.}, Northeastern University
\AND
\Name{Lea Genuit} \Email{lea@fiddler.ai}\\
\addr Fiddler Labs
\AND
\Name{Mary Reagan} \Email{mary@fiddler.ai}\\
\addr Fiddler Labs
}

\begin{document}

\maketitle

\begin{abstract}
Machine Learning or Artificial Intelligence algorithms have gained considerable scrutiny in recent times owing to their propensity towards imitating and amplifying existing prejudices in society. This has led to a niche but growing body of work that identifies and attempts to fix these biases. A first step towards making these algorithms more fair is designing metrics that measure unfairness. Most existing work in this field deals with either a binary view of fairness (protected vs. unprotected groups) or politically defined categories (race or gender). Such categorization misses the important nuance of intersectionality - biases can often be amplified in subgroups that combine membership from different categories, especially if such a subgroup is particularly underrepresented in historical platforms of opportunity.

In this paper, we discuss why fairness metrics need to be looked at under the lens of intersectionality, identify existing work in intersectional fairness, suggest a simple worst case comparison method to expand the definitions of existing group fairness metrics to incorporate intersectionality, and finally conclude with the social, legal and political framework to handle intersectional fairness in the modern context.
\end{abstract}
\begin{keywords}
intersectionality, fair machine learning, social justice, ethical artificial intelligence
\end{keywords}

\section{Introduction}

The use of machine learning algorithms is ubiquitous in the developed world. It has become an integral part of society, affecting the lives of millions of people. Algorithmic decisions vary from low-stakes determinations, like product or film recommendations, to high-impact like loan or credit approval \cite{mukerjee2002multi}, hiring recommendations \cite{bogen2018help}, facial recognition \cite{vasilescu2002multilinear} and prison recidivism \cite{corbett2018measure}. With this direct impact on people's lives, the need for fair and unbiased algorithms is paramount. It is critical that algorithms do not replicate and enhance existing societal biases, including those rooted in differences of race, gender, or sexual orientation.


To tackle these problems, both fairness and bias need to be clearly defined. Currently, there does not exist a single universally agreed upon definition of fairness. Anti-discrimination legislation exists in various jurisdictions around the world. In the US, anti-discrimination laws exist under the Civil Rights Act \cite{berg1964equal}, and under specific areas like credit lending \footnote{\url{https://www.justice.gov/crt/equal-credit-opportunity-act-3}} and housing\footnote{\url{https://www.justice.gov/crt/fair-housing-act-1}}. There have also been efforts to introduce legislation combating algorithmic bias\footnote{\url{https://www.congress.gov/bill/116th-congress/house-bill/2231/all-info}}. In the European Union, the General Data Protection Regulation (GDPR) provides for regulations regarding digital profiling, data collection, and a \say{right to explanation} \cite{goodman2017european}. Under Indian law, quotas for \textit{scheduled castes}, \textit{scheduled tribes} and \textit{other backward classes} are mandated in public education and government employment.\footnote{\url{http://www.legalservicesindia.com/article/1145/Reservations-In-India.html}}

We begin with the broad definition of fairness as \say{the absence of prejudice or preference for an individual or group based on their characteristics}. Bias can also exist in a variety of forms. \citet{mehrabi2019fairness} provides an excellent overview on the differing types of bias and discrimination. In general, a fair machine learning algorithm is one that does not favor or make prejudice towards an individual or a group.

While most early fairness research focused on binary fairness metrics (protected vs. unprotected groups), newer methods to address fairness have begun to incorporate intersectional frameworks. These frameworks are derived from the third wave of feminist thought, which is rooted in the understanding of the interconnected nature of social categories, like race, gender, sexual orientation, and class \cite{crenshaw1989demarginalizing}. The intersection of these categories creates differing levels of privilege or disadvantage for the various possible subgroups. There exist legal precedents for discrimination under an intersectional lens : The Equal Employment Opportunity Commission (EEOC) describes some Intersectional Discrimination/Harassment examples\footnote{\urlx{https://www.eeoc.gov/initiatives/e-race/significant-eeoc-racecolor-casescovering-private-and-federal-sectors\#intersectional}}. \citet{buolamwini2018gender} examined gender classification algorithms for facial image data and found that they performed substantially better on male faces than female faces. However, the largest performance drops came when both race and gender were considered, with darker skinned women disproportionately affected having a misclassification rate of $\approx$30\%.

\begin{figure}[h]
\centering
\includegraphics[width=0.6\columnwidth]{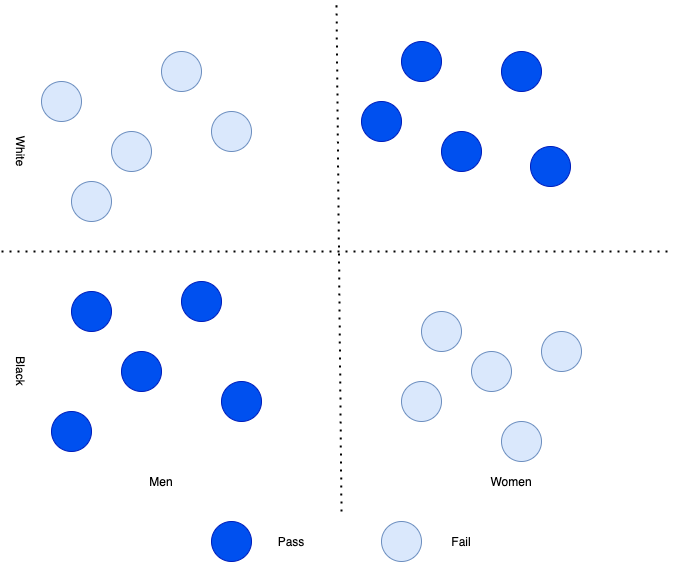} 
\caption{An example of ``fairness gerrymandering"}
\label{fig:intsec}
\end{figure}

The example in Figure \ref{fig:intsec} describes the importance of intersectional fairness. In the figure, we observe equal numbers of black and white people pass. Similarly, there is an equal number of men and women passing. However, this classification is unfair because we don’t have any black women and white men that passed, and all black men and white women passed. We observe the bias only while looking at the subgroups when we take race and gender as protected attributes. This phenomenon was called \textit{\say{Fairness Gerrymandering}} by \citet{kearns2018preventing}.

Additionally, there are minorities that have historically faced discrimination around the world, but due to their sparse population, empirical evidence of discrimination against them is difficult to trace, for example, the indigenous population \cite{paradies2006systematic,king2009indigenous}, or trans people \cite{feldman2016priorities,reisner2016advancing, bockting2016adult}. This causes machine learning practitioners to either disinclude these groups from their training datasets due to statistical insignificance, or worse, conflate them with other minorities to create a general \say{protected} category, which leads to the same sort of neglected bias as shown in figure \ref{fig:intsec}.

In this paper, we discuss the notion of intersectional group fairness. After introducing existing related work, we define a combinatorial approach giving subsets of the population. With this definition of subgroups, we introduce a measure of the worst case disparity using existing fairness metrics, to discover biases against underserved subgroups. We then show how this method can be applied to classification models, ranking models, and models with continuous output. We end the paper with a discussion about the limitations of our approach and future work.

\section{Related Work}

\subsection{Individual and Group Fairness}
Fair machine learning differentiates \textit{group} and \textit{individual} fairness measures. While group fairness metrics focus on treating two different groups equally, individual fairness metrics focus on treating similar individuals similarly. \citet{binns2019ind-group} introduces those two notions and discusses the motivations behind individual and group fairness. In this paper, we focus on group fairness metrics.

\subsection{Binary fairness metrics} A large majority of research in algorithmic fairness has covered fairness metrics for a single protected attribute \cite{corbett2018measure}. \citet{hardt2016equality} introduces the definitions of Equalized odds and equal opportunity, two measures for discrimination against a binary sensitive attribute. \citet{Verma2018binclass} collected some known binary fairness metrics for classification models and demonstrated each metric with a unique example on the German credit dataset. In their example, the protected class is \textit{Gender}, which has two values \textit{female} and \textit{male}.

\subsection{Intersectional Fairness} More recently, however, some work has begun to address the issue of intersectionality in AI by providing statistical frameworks that control for bias within multiple subgroups. \citet{hebert2018multicalibration} introduces the idea of \textit{multi-calibration} which gives meaningful predictions for overlapping subgroups in a larger protected group. \citet{kearns2018preventing} developed an analogous method named \textit{rich subgroup fairness} for false positive and negative constraints that hold over an infinitely large collection of subgroups. \citet{kim2019multiaccuracy} extend these methods for classifiers to be equally accurate on a combinatorially large collection of all subgroups. \citet{mary2019fairness} present the Renyi correlation coefficient as a fairness metric for datasets with continuous protected attributes. Finally, \citet{foulds2020intersectional} introduce \textit{differential fairness} (DF), as an intersectional fairness metric.

\section{Intersectional group fairness metrics}

In this section, we discuss our intersectional fairness metrics framework. We outline our definition of a subgroup of the population, define a \textit{worst case} disparity metric that we call the \textit{min-max ratio} and describe how we can operationalize the notion of \textit{min-max ratio} to encompass intersectionality in existing metrics of fairness.

\subsection{Subgroup definition}
For the purposes of this paper, similar to \citet{kearns2018preventing}, we define a subgroup $sg_{a_1....a_n}$ as a set containing the intersection of all members who belong to groups $g_{a_1}$ through $g_{a_n}$, where $a_1$, $a_2$...$a_n$ are marginal protected attributes, like race, gender, etc. Formally,

\begin{equation}
    sg_{a_1 \times a_2 \times ... \times a_n} = g_{a_1} \cap g_{a_2} ... \cap g_{a_n}
\end{equation}

Hence, for example, if $g_1$(race) $\in$ \{black, white\} and $g_2$(gender) $\in$ \{man, woman\}, then $sg \in$ \{black women, black men, white women, white men\} and N = $|sg| = 2 \times 2 = 4$. 

This combinatorial, or cartesian product of attributes approach gives us subsets of the original dataset, where in each subgroup, the members have all the protected attributes of the groups they were composed of.

\subsection{Worst Case Disparity} \label{worstcase}

We introduce a simple concept to measure the worst case disparity using existing fairness metrics to incorporate intersectionality - the \textit{min-max ratio}. In the vein of \cite{rawls2001justice} principle for distributive justice, the idea essentially is to measure the value of the given fairness metric for every subgroup $sg_{i}$ then take the ratio of the \textit{minimum} and  \textit{maximum} values from this given list. The further this ratio is from 1, the greater the disparity is between subgroups. \textbf{This \textit{min-max ratio} technique allows us to encompass the entire breadth of possible subgroups in a dataset, by considering the worst case scenario in terms of adverse impact.} For fairness metrics that are already comparative ratios of two groups, we redefine it by calculating the said ratio for all possible permutations of two subgroups and then simply take the minimum, also the worst possible case.

We discuss some of the most commonly used metrics in the literature below and show how we use the \textit{worst possible case} framing to incorporate intersectionality.

\subsection{Fair Classification metrics}
Several fair classification metrics exist in literature. We discuss four group fairness metrics below from \citet{mehrabi2019fairness} and \citet{gbenefit}.

\subsubsection{Demographic parity} 
According to demographic parity, the proportion of each segment of a protected class should receive positive outcomes at equal rates. Mathematically, demographic parity compares the pass rate (rate of positive outcome) of two groups. Demographic parity is satisfied for a predictor $\hat{Y}$ and for a member $A$ if:

\begin{equation}
P(\hat{Y}|A \in sg_i) = P(\hat{Y}|A \in sg_j); \forall i, j \in N, i \neq j
\end{equation}

where N is the total number of subgroups. Demographic parity is also known as statistical parity \cite{dwork2012fairness,kusner2017counterfactual}.

Using our \textit{worst case, min-max ratio} definition, \textit{Demographic parity ratio} (DPR) would be defined as:

\begin{equation}
    \text{DPR} = \frac{\text{min}\{P(\hat{Y}|A \in sg_i) \forall i \in N \}}{\text{max}\{P(\hat{Y}|A \in sg_i) \forall i \in N \}}
\end{equation} \label{eq:dpr}

Disparate impact, as defined under the guideline by the Equal Employment Opportunity \citet{equal1979adoption} is similar to the demographic parity metric. It is intended as a way to measure indirect and unintentional discrimination in which certain decisions disproportionately affect members of a protected group. Disparate impact compares the pass rate of one group versus another. The Four-Fifths rule states that the ratio of the pass rate of group 1 to the pass rate of group 2 has to be greater than 80\% (groups 1 and 2 interchangeable). Using our worst case definition, intersectional disparate impact (DI) is defined as the minimum disparate impact between all possible pairs of subgroups $sg$.

\begin{equation}
    \text{DI} = \text{min} \left\{ \frac{P(\hat{Y}|A \in sg_i)}{P(\hat{Y}|A \in sg_j)};  \forall i, j \in N, i \neq j \right\}
\end{equation}

\subsubsection{Conditional statistical parity}
Conditional statistical parity extends demographic parity by permitting a set of legitimate attributes to affect the outcome \cite{corbett2017algorithmic}. Conditional statistical parity is satisfied for a predictor $\hat{Y}$, a member $A$ with a set of legitimate attributes $L$ if:

\begin{equation}
    P(\hat{Y}|L=1, A \in sg_i) = P(\hat{Y}|L=1, A \in sg_j)
    \forall i, j \in N, i \neq j
\end{equation}

Using the \textit{worst case, min-max ratio} definition, \textit{Conditional statistical parity ratio} (CSPR) would be defined just like equation \ref{eq:dpr}:

\begin{equation}
    \text{CSPR} = \frac{\text{min}\{P(\hat{Y}|L=1, A \in sg_i) \forall i \in N \}}{\text{max}\{P(\hat{Y}|L=1, A \in sg_i) \forall i \in N \}}
\end{equation} \label{eq:cspr}

\subsubsection{Equal opportunity}
Equal opportunity or True Positive Rate Parity states that all members should be treated equally or similarly and not disadvantaged by prejudice or bias. Mathematically, it compares True Positive Rate (TPR) of the classifier between the protected group and the unprotected group\footnote{TPR is the probability that a ground truth positive observation is correctly classified as positive.} \cite{hardt2016equality}. Equal opportunity for a binary predictor $\hat{Y}$ and a member $A$, is satisfied if:

\begin{equation}
    P(\hat{Y}=1|A \in sg_i, Y=1) = P(\hat{Y}=1|A \in sg_j , Y=1)
    \forall i, j \in N, i \neq j 
\end{equation}

Using the \textit{worst case, min-max ratio} definition, \textit{Equal opportunity ratio} (EOppR) would be defined as:

\begin{equation}
     \text{EOppR} =\frac{\text{min}\{P(\hat{Y}=1|A \in sg_i, Y=1) \forall i \in N \}}{\text{max}\{P(\hat{Y}=1|A \in sg_i, Y=1) \forall i \in N \}}
\end{equation}

In a similar vein, there can exist \textit{True Negative Rate Parity}, \textit{False Positive Rate Parity} and \textit{False Negative Rate Parity}. \cite{hardt2016equality} propose \textit{Equalized Odds} as a method to generalize the Equal Opportunity metric by comparing all these different parities. 






\subsubsection{Group Benefit Equality}

Group Benefit Equality, introduced by \cite{gbenefit} aims to be useful in the domain of healthcare. Group benefit equality measures the predicted rate of passing for a subgroup compared to the actual rate of passing. Mathematically, this is defined as:

\begin{equation}
\begin{split}
    P(\hat{Y}|A \in sg_i) = P(Y | A \in sg_i)
    \forall i \in N
\end{split}
\end{equation}

And, Group benefit ratio for a subgroup is defined as:

\begin{equation}
    \text{GBR}_{sg_i} = \frac{P(\hat{Y}|A \in sg_i)}{P(Y | A \in sg_i)}
    ; \forall i \in N
\end{equation}

Using the \textit{worst case, min-max ratio} definition, \textit{Group benefit ratio} (GBR\_INT) would be defined intersectionally as:

\begin{equation}
    \text{GBR\_INT} = \frac{\text{min}\{ \text{GBR}_{sg_i},\forall i \in N
     \}}{\text{max}\{ \text{GBR}_{sg_i},\forall i \in N
     \}}
\end{equation}

\subsection{Multi-class classification models}

For multiclass classification models, we present a modified version of the Equalized Odds metric, except instead of a binary \textit{positive} or \textit{negative} label, we measure the odds ratio for each possible discrete output, and then take the worst odds ratio among all outputs.

For instance, if a multiclass classifier has five possible output classes, we calculate the min-max ratio for each output class $y$, and then take the minimum of those five values as our final metric, since it is the \textit{worst possible scenario.} Formally, Multiclass Equalized Odds Ratio (M-EOddR) is defined as:

\begin{equation}
     \text{M-EOddR} = \text{min}\left\{\frac{\text{min}\{P(\hat{Y}=y_k|A \in sg_i), \forall i \in N \}}{\text{max}\{P(\hat{Y}=y_k|A \in sg_i), \forall i \in N\}} \right\} 
    \forall k\in K
\end{equation}

where K is the set of all possible output classes. The closer the value of M-EOddR is to 1, the lower the disparity is of the classifier's performance among the various subgroups for all possible output classes.

\begin{figure*}[h]
    \includegraphics[width=\columnwidth]{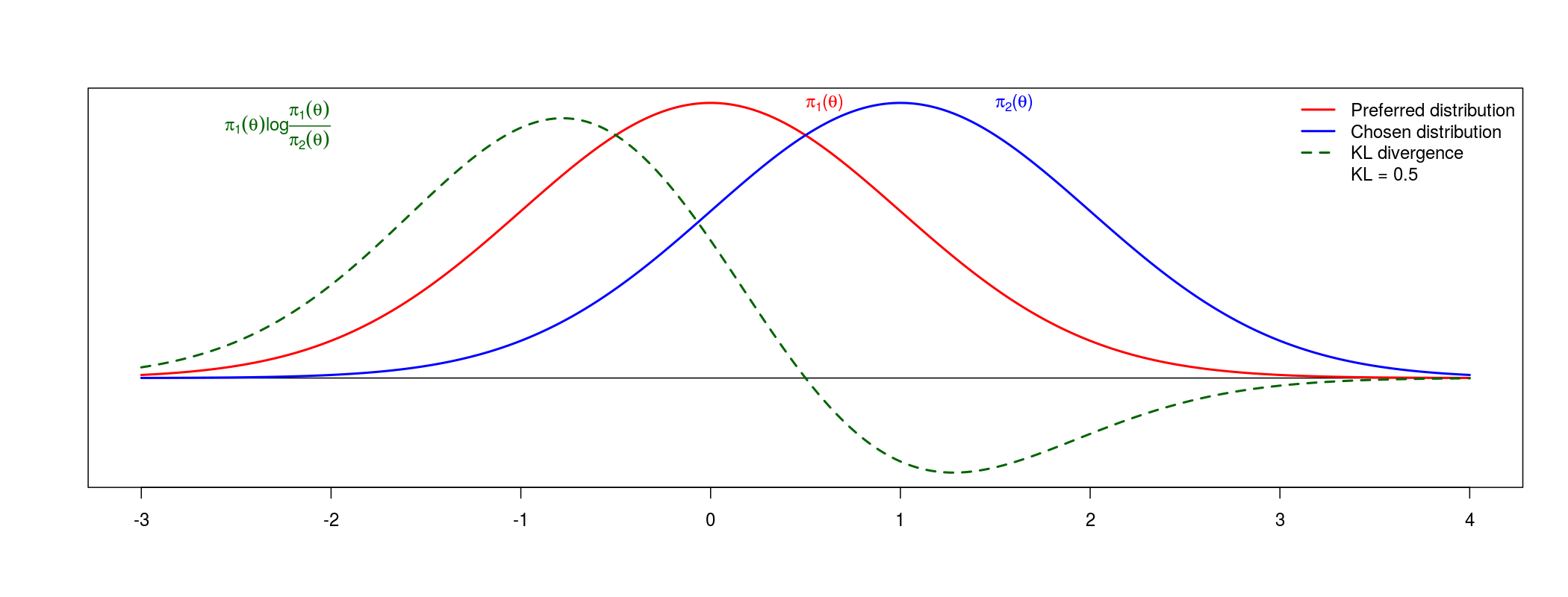}
           \caption{KL divergence example between two distributions adapted from \citet{veen2018using}. In this example $\pi_1$ is a standard normal distribution and $\pi_2$ is a normal distribution with a mean of 1 and a variance of 1. The value of the KL divergence is equal to the area under the curve of the function (green line). The area under the green line above the x-axis adds to the divergence, while the area under the x-axis subtracts from the divergence.} 
\end{figure*}

\subsection{Models with continuous output}

We can extend the worst possible case framing for models which produce a continuous output, like regression models, or recommendation models that provide relevance scores. The Kullback-Leibler (KL) divergence\footnote{Here we use KL Divergence as our base metric, although this method would work for any distribution comparison metric.} between two distributions $q$ and $p$ is defined as the following:
\begin{equation}
    D_{KL}(\pi_1||\pi_2) = \int_{\infty}^{\infty} \pi_1(x) log(\frac{\pi_1(x)}{\pi_2(x)}) dx
\end{equation}
In the context of intersectional fairness, we compute the KL divergence between the model output distributions of all possible pairs of two subgroups, and we display the maximum KL divergence value obtained, since it is the \textit{worst case scenario}. If this value is close to 0, the two subgroups have similar distributions, as well as the other subgroups.

Thus, Worst case KL Divergence (W-$D_{KL}$) is formally defined as:

\begin{equation}
    \text{W-}D_{KL} = \text{max} \left\{  \int_{\infty}^{\infty} \pi_{sg_i}(x) log(\frac{\pi_{sg_i}(x)}{\pi_{sg_j}(x)} \,dx  \right\}
    \forall i,j \in N, i \neq j
\end{equation}

\subsection{Ranking metrics}

Existing fair ranking metrics in the literature can be divided broadly into two classes - representation based \cite{yang2017measuring} and exposure based \cite{singh2018fairness, sapiezynski2019quantifying}. We pick one of each kind and redefine them under the light of intersectionality.
\subsubsection{Skew}

The \textit{representation-based} metric we discuss is skew@k \cite{geyik2019fairness}. For a ranked list $\tau$, the $\text{Skew}$ for subgroup $sg_i$ at the top $k$ is defined as 
\begin{equation}
    \text{Skew}_{sg_{i}}@k(\tau)= \frac{p_{\tau^{k},sg_{i}}}{p_{q,sg_{i}}}
\end{equation}
where $p_{\tau^{k},sg_{i}}$ represents the fraction of members from subgroup $sg_{i}$ among the top $k$ items in $\tau$, and $p_{q,g_{i}}$ represents the fraction of members from to subgroup $sg_{i}$ in the overall population $q$. Ideally, $\text{Skew}_{g_{i}}@k$ should be close to one for each $sg_{i}$ and $k$, to show that people from $sg_{i}$ are represented in $\tau$ proportionally relative to the overall population.

Using our \textit{worst case} method, the skew ratio at K (SR@K) is defined as:

\begin{equation}
     \text{SR@K} = \frac{
     \text{min}\{\text{Skew}_{sg_{i}}@k(\tau),\forall i \in N\}
     }{
     \text{max}\{\text{Skew}_{sg_{i}}@k(\tau),\forall i \in N\}
     }
\end{equation}

\begin{figure*}[t]
\centerfloat
    \includegraphics[width=1.2\textwidth]{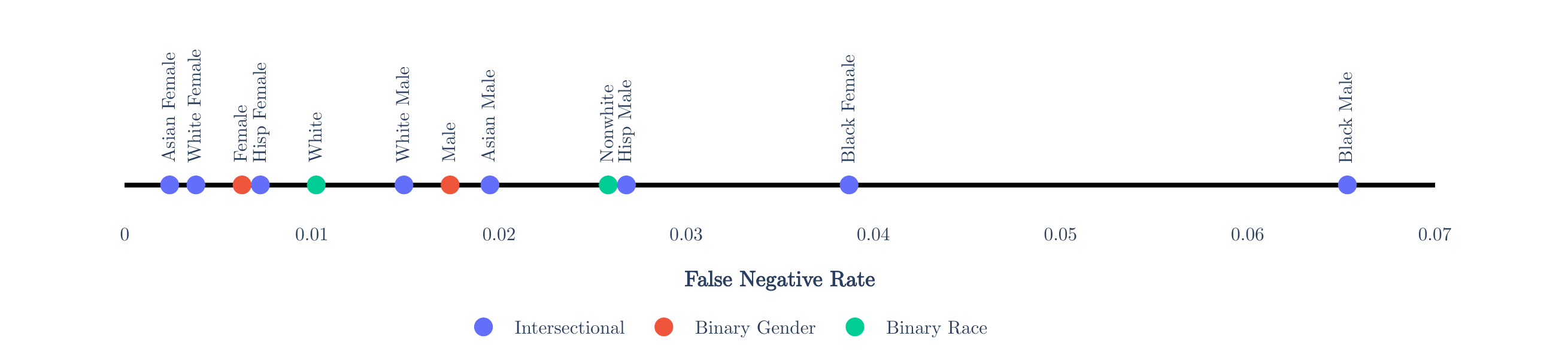}
          \caption{The different values for the False Negative Rate Parity measurement for the LSAC case study. Minority subgroups under intersectionality show a far greater range of disparity than what the binary metrics would suggest.} 
\end{figure*}

\subsubsection{Attention}

Attention is the \textit{exposure-based} metric we discuss here. Ranking problems are unique from classification problems in the sense that the position of a ranked item, even within the top K results, can draw significantly different levels of visual attention. Previous research shows that people's attention sharply drops off after the first few items in a ranked list \cite{mullick2019public}. Different papers have modeled visual attention as a function of the position K as a logarithmic distribution \cite{singh2018fairness}, a geometric distribution, or other sharply falling distributions with increasing rank \cite{sapiezynski2019quantifying}. Assuming the attention distribution function of an item to be Att(k), the mean attention per subgroup is defined as:

\begin{equation}
    \text{MA}_{sg_i} = \frac{1}{|sg_{i}|}\sum_{k=1}^{|\tau|}\text{Att(k)} ~\text{where}~ sg_k^{\tau} = sg_{i}
\end{equation}

And, using our \textit{worst case} method, the attention ratio (AR) is defined as:

\begin{equation}
    \text{AR} = \frac{\text{min}\{ \text{MA}_{sg_i},\forall i \in N
     \}}{\text{max}\{ \text{MA}_{sg_i},\forall i \in N
     \}}
\end{equation}

\section{Case Study and LSAC Dataset}
As an example application of the framework described in this paper, we present a case study on a trained tensorflow model\footnote{\url{https://github.com/tensorflow/fairness-indicators/blob/master/g3doc/tutorials/Fairness_Indicators_Pandas_Case_Study.ipynb}} outputs on the Law School Admissions Council (LSAC) dataset\footnote{\url{https://eric.ed.gov/?id=ED469370}}, where the classifier predicts whether a candidate passed the bar exam. We calculated metrics for binary values where only race (Table 1) or only gender (Table 2) is examined for the false negative rates (FNR) and also for an intersectional FNR metric, where both race and gender are used (Table 3). 
\begin{table}[h]
\centering
\begin{tabular}{|c | c|} 
 \hline
\textbf{Race} & \textbf{FNR}\\[0.5ex] 
 \hline
nonwhite & 0.025829\\[0.5ex] 
white & 0.010230\\[0.5ex] 
 \hline
\end{tabular}
\caption{False negative rates using race}
\label{table:1}
\end{table}

\begin{table}[h]
\centering
\begin{tabular}{|c | c|} 
 \hline
\textbf{Gender} & \textbf{FNR}\\[0.5ex] 
 \hline
woman & 0.006267\\[0.5ex] 
man & 0.017384\\[0.5ex] 
 \hline
\end{tabular}
\caption{False negative rates using gender}
\label{table:2}
\end{table}

\begin{table}[h]
\centering
\begin{tabular}{|c |c | c|} 
 \hline
 \textbf{Gender} & \textbf{Race} & \textbf{FNR} \\ [0.5ex] 
 \hline
 woman &	asian	 & \textbf{0.002398} \\ [0.5ex] 
woman &	black &	0.038700 \\ [0.5ex] 
woman &	hisp	& 0.007246 \\[0.5ex] 
woman & white & 0.003802 \\[0.5ex] 
man & asian & 0.019512 \\[0.5ex] 
man & black & \textbf{0.065327} \\[0.5ex] 
man & hisp & 0.026804 \\[0.5ex] 
man & white & 0.014920 \\[0.5ex] 
 \hline
\end{tabular}
\caption{False negative rates using intersectional race and gender subgroups.}
\label{table:3}
\end{table}

The tables show clear examples of how viewing fairness metrics for only one group or protected class can obscure inequality for combined subgroups. When using only the gender lens, the FNR is lower for women, $\approx{0.006}$ when compared to men, $\approx{ 0.017}$ (Table 2). However the trend reverses for certain subgroups when race is added. For example with Black women having higher FNR at $\approx{ 0.039}$ when compared either to white men at $\approx{ 0.015}$ or asian men at $\approx{ 0.020}$ (Table 3).  The min/max ratio is 0.002398/0.065327 = 0.036, which is far from the ideal value of 1. The model therefore fails to achieve intersectional fairness under FNR parity.

\section{Discussion}

\subsection{Conclusion} In this paper, we introduce the \textit{worst-case} comparison as a simple, easily comprehensible method to surface hidden biases that commonly used fairness metrics may not be able to show. We establish the importance of introducing such modifications to better serve minorities with sparse populations and show how the method can be applied to a diverse range of model metrics, thereby being easy for practitioners and researchers to adapt without significantly changing their existing fairness monitoring systems.

\subsection{Limitations and Future Work} \label{sec:limitation}
The idea of creating combinatorial subgroups has a couple of caveats: It does not take into account partial group membership (for instance, a person who identifies as multiracial), or continuous variables (for example, instead of treating age as an integer, we would convert the age attribute as discrete buckets). We encourage researchers to expand our method to include partial group membership and continuous attributes. 

Secondly, creating a combinatorially large number of subgroups inevitably leads to subgroups which have a very small number of members, thereby demonstrating the effects of Simpson's Paradox \cite{blyth1972simpson}. A possible direction of research could be to introduce statistical significance measures for such small subgroups, and suggest thumb rules for subgroup creation via empirical measurements.

\acks{
The authors would like to thank Joshua Rubin, Amit Paka, Jack Reidy, Aalok Shanbhag, Christo Wilson, and the anonymous reviewers for their helpful discussions and suggestions.}

\bibliography{ref}

\begin{thebibliography}{35}
\providecommand{\natexlab}[1]{#1}
\providecommand{\url}[1]{\texttt{#1}}
\expandafter\ifx\csname urlstyle\endcsname\relax
  \providecommand{\doi}[1]{doi: #1}\else
  \providecommand{\doi}{doi: \begingroup \urlstyle{rm}\Url}\fi

\bibitem[Berg(1964)]{berg1964equal}
Richard~K Berg.
\newblock Equal employment opportunity under the civil rights act of 1964.
\newblock \emph{Brook. L. Rev.}, 31:\penalty0 62, 1964.

\bibitem[Binns(2019)]{binns2019ind-group}
Reuben Binns.
\newblock On the apparent conflict between individual and group fairness.
\newblock \emph{CoRR}, abs/1912.06883, 2019.
\newblock URL \url{http://arxiv.org/abs/1912.06883}.

\bibitem[Blyth(1972)]{blyth1972simpson}
Colin~R Blyth.
\newblock On simpson's paradox and the sure-thing principle.
\newblock \emph{Journal of the American Statistical Association}, 67\penalty0
  (338):\penalty0 364--366, 1972.

\bibitem[Bockting et~al.(2016)Bockting, Coleman, Deutsch, Guillamon, Meyer,
  Meyer~III, Reisner, Sevelius, and Ettner]{bockting2016adult}
Walter Bockting, Eli Coleman, Madeline~B Deutsch, Antonio Guillamon, Ilan
  Meyer, Walter Meyer~III, Sari Reisner, Jae Sevelius, and Randi Ettner.
\newblock Adult development and quality of life of transgender and gender
  nonconforming people.
\newblock \emph{Current opinion in endocrinology, diabetes, and obesity},
  23\penalty0 (2):\penalty0 188, 2016.

\bibitem[Bogen and Rieke(2018)]{bogen2018help}
Miranda Bogen and Aaron Rieke.
\newblock Help wanted: An examination of hiring algorithms, equity, and bias.
\newblock 2018.

\bibitem[Buolamwini and Gebru(2018)]{buolamwini2018gender}
Joy Buolamwini and Timnit Gebru.
\newblock Gender shades: Intersectional accuracy disparities in commercial
  gender classification.
\newblock In \emph{Conference on fairness, accountability and transparency},
  pages 77--91, 2018.

\bibitem[Commission et~al.(1979)]{equal1979adoption}
Equal Employment~Opportunity Commission et~al.
\newblock Adoption of questions and answers to clarify and provide a common
  interpretation of the uniform guidelines on employee selection procedures.
\newblock \emph{Federal Register}, 44\penalty0 (43):\penalty0 11996--12009,
  1979.

\bibitem[Corbett-Davies and Goel(2018)]{corbett2018measure}
Sam Corbett-Davies and Sharad Goel.
\newblock The measure and mismeasure of fairness: A critical review of fair
  machine learning.
\newblock \emph{arXiv preprint arXiv:1808.00023}, 2018.

\bibitem[Corbett-Davies et~al.(2017)Corbett-Davies, Pierson, Feller, Goel, and
  Huq]{corbett2017algorithmic}
Sam Corbett-Davies, Emma Pierson, Avi Feller, Sharad Goel, and Aziz Huq.
\newblock Algorithmic decision making and the cost of fairness.
\newblock In \emph{Proceedings of the 23rd acm sigkdd international conference
  on knowledge discovery and data mining}, pages 797--806, 2017.

\bibitem[Crenshaw(1989)]{crenshaw1989demarginalizing}
Kimberl{\'e} Crenshaw.
\newblock Demarginalizing the intersection of race and sex: A black feminist
  critique of antidiscrimination doctrine, feminist theory and antiracist
  politics.
\newblock \emph{u. Chi. Legal f.}, page 139, 1989.

\bibitem[Dwork et~al.(2012)Dwork, Hardt, Pitassi, Reingold, and
  Zemel]{dwork2012fairness}
Cynthia Dwork, Moritz Hardt, Toniann Pitassi, Omer Reingold, and Richard Zemel.
\newblock Fairness through awareness.
\newblock In \emph{Proceedings of the 3rd innovations in theoretical computer
  science conference}, pages 214--226, 2012.

\bibitem[Feldman et~al.(2016)Feldman, Brown, Deutsch, Hembree, Meyer,
  Meyer-Bahlburg, Tangpricha, T’Sjoen, and Safer]{feldman2016priorities}
Jamie Feldman, George~R Brown, Madeline~B Deutsch, Wylie Hembree, Walter Meyer,
  Heino~FL Meyer-Bahlburg, Vin Tangpricha, Guy T’Sjoen, and Joshua~D Safer.
\newblock Priorities for transgender medical and health care research.
\newblock \emph{Current opinion in endocrinology, diabetes, and obesity},
  23\penalty0 (2):\penalty0 180, 2016.

\bibitem[Foulds et~al.(2020)Foulds, Islam, Keya, and
  Pan]{foulds2020intersectional}
James~R Foulds, Rashidul Islam, Kamrun~Naher Keya, and Shimei Pan.
\newblock An intersectional definition of fairness.
\newblock In \emph{2020 IEEE 36th International Conference on Data Engineering
  (ICDE)}, pages 1918--1921. IEEE, 2020.

\bibitem[Gartner(2020)]{gbenefit}
Joseph Gartner.
\newblock {A New Metric for Quantifying Machine Learning Fairness in
  Healthcare}.
\newblock
  \urlx{https://closedloop.ai/a-new-metric-for-quantifying-fairness-in-healthcare/},
  2020.

\bibitem[Geyik et~al.(2019)Geyik, Ambler, and Kenthapadi]{geyik2019fairness}
Sahin~Cem Geyik, Stuart Ambler, and Krishnaram Kenthapadi.
\newblock Fairness-aware ranking in search \& recommendation systems with
  application to linkedin talent search.
\newblock In \emph{Proceedings of the 25th ACM SIGKDD International Conference
  on Knowledge Discovery \& Data Mining}, pages 2221--2231, 2019.

\bibitem[Goodman and Flaxman(2017)]{goodman2017european}
Bryce Goodman and Seth Flaxman.
\newblock European union regulations on algorithmic decision-making and a
  “right to explanation”.
\newblock \emph{AI magazine}, 38\penalty0 (3):\penalty0 50--57, 2017.

\bibitem[Hardt et~al.(2016)Hardt, Price, and Srebro]{hardt2016equality}
Moritz Hardt, Eric Price, and Nati Srebro.
\newblock Equality of opportunity in supervised learning.
\newblock In \emph{Advances in neural information processing systems}, pages
  3315--3323, 2016.

\bibitem[H{\'e}bert-Johnson et~al.(2018)H{\'e}bert-Johnson, Kim, Reingold, and
  Rothblum]{hebert2018multicalibration}
{\'U}rsula H{\'e}bert-Johnson, Michael Kim, Omer Reingold, and Guy Rothblum.
\newblock Multicalibration: Calibration for the (computationally-identifiable)
  masses.
\newblock In \emph{International Conference on Machine Learning}, pages
  1939--1948, 2018.

\bibitem[Kearns et~al.(2018)Kearns, Neel, Roth, and Wu]{kearns2018preventing}
Michael Kearns, Seth Neel, Aaron Roth, and Zhiwei~Steven Wu.
\newblock Preventing fairness gerrymandering: Auditing and learning for
  subgroup fairness.
\newblock In \emph{International Conference on Machine Learning}, pages
  2564--2572. PMLR, 2018.

\bibitem[Kim et~al.(2019)Kim, Ghorbani, and Zou]{kim2019multiaccuracy}
Michael~P Kim, Amirata Ghorbani, and James Zou.
\newblock Multiaccuracy: Black-box post-processing for fairness in
  classification.
\newblock In \emph{Proceedings of the 2019 AAAI/ACM Conference on AI, Ethics,
  and Society}, pages 247--254, 2019.

\bibitem[King et~al.(2009)King, Smith, and Gracey]{king2009indigenous}
Malcolm King, Alexandra Smith, and Michael Gracey.
\newblock Indigenous health part 2: the underlying causes of the health gap.
\newblock \emph{The lancet}, 374\penalty0 (9683):\penalty0 76--85, 2009.

\bibitem[Kusner et~al.(2017)Kusner, Loftus, Russell, and
  Silva]{kusner2017counterfactual}
Matt~J Kusner, Joshua Loftus, Chris Russell, and Ricardo Silva.
\newblock Counterfactual fairness.
\newblock In \emph{Advances in neural information processing systems}, pages
  4066--4076, 2017.

\bibitem[Mary et~al.(2019)Mary, Calauz{\`e}nes, and
  El~Karoui]{mary2019fairness}
J{\'e}r{\'e}mie Mary, Cl{\'e}ment Calauz{\`e}nes, and Noureddine El~Karoui.
\newblock Fairness-aware learning for continuous attributes and treatments.
\newblock In \emph{International Conference on Machine Learning}, pages
  4382--4391, 2019.

\bibitem[Mehrabi et~al.(2019)Mehrabi, Morstatter, Saxena, Lerman, and
  Galstyan]{mehrabi2019fairness}
Ninareh Mehrabi, Fred Morstatter, Nripsuta Saxena, Kristina Lerman, and Aram
  Galstyan.
\newblock A survey on bias and fairness in machine learning.
\newblock \emph{CoRR}, abs/1908.09635, 2019.
\newblock URL \url{http://arxiv.org/abs/1908.09635}.

\bibitem[Mukerjee et~al.(2002)Mukerjee, Biswas, Deb, and
  Mathur]{mukerjee2002multi}
Amitabha Mukerjee, Rita Biswas, Kalyanmoy Deb, and Amrit~P Mathur.
\newblock Multi--objective evolutionary algorithms for the risk--return
  trade--off in bank loan management.
\newblock \emph{International Transactions in operational research}, 9\penalty0
  (5):\penalty0 583--597, 2002.

\bibitem[Mullick et~al.(2019)Mullick, Ghosh, Dutt, Ghosh, and
  Chakraborty]{mullick2019public}
Ankan Mullick, Sayan Ghosh, Ritam Dutt, Avijit Ghosh, and Abhijnan Chakraborty.
\newblock Public sphere 2.0: Targeted commenting in online news media.
\newblock In \emph{European Conference on Information Retrieval}, pages
  180--187. Springer, 2019.

\bibitem[Paradies(2006)]{paradies2006systematic}
Yin Paradies.
\newblock A systematic review of empirical research on self-reported racism and
  health.
\newblock \emph{International journal of epidemiology}, 35\penalty0
  (4):\penalty0 888--901, 2006.

\bibitem[Rawls(2001)]{rawls2001justice}
John Rawls.
\newblock \emph{Justice as fairness: A restatement}.
\newblock Harvard University Press, 2001.

\bibitem[Reisner et~al.(2016)Reisner, Deutsch, Bhasin, Bockting, Brown,
  Feldman, Garofalo, Kreukels, Radix, Safer, et~al.]{reisner2016advancing}
Sari~L Reisner, Madeline~B Deutsch, Shalender Bhasin, Walter Bockting, George~R
  Brown, Jamie Feldman, Rob Garofalo, Baudewijntje Kreukels, Asa Radix,
  Joshua~D Safer, et~al.
\newblock Advancing methods for us transgender health research.
\newblock \emph{Current opinion in endocrinology, diabetes, and obesity},
  23\penalty0 (2):\penalty0 198, 2016.

\bibitem[Sapiezynski et~al.(2019)Sapiezynski, Zeng, E~Robertson, Mislove, and
  Wilson]{sapiezynski2019quantifying}
Piotr Sapiezynski, Wesley Zeng, Ronald E~Robertson, Alan Mislove, and Christo
  Wilson.
\newblock Quantifying the impact of user attentionon fair group representation
  in ranked lists.
\newblock In \emph{Companion Proceedings of The 2019 World Wide Web
  Conference}, pages 553--562, 2019.

\bibitem[Singh and Joachims(2018)]{singh2018fairness}
Ashudeep Singh and Thorsten Joachims.
\newblock Fairness of exposure in rankings.
\newblock In \emph{Proceedings of the 24th ACM SIGKDD International Conference
  on Knowledge Discovery \& Data Mining}, pages 2219--2228, 2018.

\bibitem[Vasilescu and Terzopoulos(2002)]{vasilescu2002multilinear}
M~Alex~O Vasilescu and Demetri Terzopoulos.
\newblock Multilinear image analysis for facial recognition.
\newblock In \emph{Object recognition supported by user interaction for service
  robots}, volume~2, pages 511--514. IEEE, 2002.

\bibitem[Veen et~al.(2018)Veen, Stoel, Schalken, Mulder, and Van~de
  Schoot]{veen2018using}
Duco Veen, Diederick Stoel, Naomi Schalken, Kees Mulder, and Rens Van~de
  Schoot.
\newblock Using the data agreement criterion to rank experts’ beliefs.
\newblock \emph{Entropy}, 20\penalty0 (8):\penalty0 592, 2018.

\bibitem[Verma and Rubin(2018)]{Verma2018binclass}
Sahil Verma and Julia Rubin.
\newblock Fairness definitions explained.
\newblock In \emph{Proceedings of the International Workshop on Software
  Fairness}, FairWare '18, page 1–7, New York, NY, USA, 2018. Association for
  Computing Machinery.
\newblock ISBN 9781450357463.
\newblock \doi{10.1145/3194770.3194776}.
\newblock URL \url{https://doi.org/10.1145/3194770.3194776}.

\bibitem[Yang and Stoyanovich(2017)]{yang2017measuring}
Ke~Yang and Julia Stoyanovich.
\newblock Measuring fairness in ranked outputs.
\newblock In \emph{Proceedings of the 29th International Conference on
  Scientific and Statistical Database Management}, pages 1--6, 2017.

\end{thebibliography}

\appendix

\end{document}